\documentclass[12pt, anon]{l4dc2024}
\usepackage[utf8]{inputenc} % allow utf-8 input
\usepackage[T1]{fontenc}    % use 8-bit T1 fonts
\usepackage{hyperref}       % hyperlinks
\usepackage{url}            % simple URL typesetting
\usepackage{booktabs}       % professional-quality tables
\usepackage{amsfonts}       % blackboard math symbols
\usepackage{nicefrac}       % compact symbols for 1/2, etc.
\usepackage{microtype}      % microtypography
\usepackage{xcolor}         % colors
 
\usepackage{bbm}
\usepackage{mathtools}
\usepackage{algorithm}
\usepackage{algorithmic}
\usepackage{multirow}
\usepackage{hhline}
\definecolor{MyBlue}{rgb}{.851,.851,.851}

\usepackage{xcolor,colortbl}
\usepackage{caption}
\usepackage{wrapfig}
\usepackage{enumitem,kantlipsum}
\usepackage{appendix}

\usepackage[english]{babel}

\usepackage{amsmath,amsfonts,bm}
\usepackage{xspace}

\def\eqref#1{equation~\ref{#1}}
\def\1{\bm{1}}

\mathchardef\mhyphen="2D

\DeclareMathAlphabet{\mathsfit}{\encodingdefault}{\sfdefault}{m}{sl}
\SetMathAlphabet{\mathsfit}{bold}{\encodingdefault}{\sfdefault}{bx}{n}

\def\Acal{{\mathcal{A}}}

\def\Mcal{{\mathcal{M}}}

\def\Pcal{{\mathcal{P}}}

\def\Scal{{\mathcal{S}}}

\newtheorem{Assumption}{Assumption}

\title[Gradient Shaping for Multi-Constraint Safe Reinforcement Learning]{Gradient Shaping for Multi-Constraint Safe Reinforcement Learning}
\usepackage{times}
\author{%
 \Name{Yihang Yao} \Email{yihangya@andrew.cmu.edu}
 \\
 \addr Carnegie Mellon University,
       Pittsburgh, PA, USA
 \AND
 \Name{Zuxin Liu} \Email{zuxinl@andrew.cmu.edu}
 \\
 \addr Carnegie Mellon University,
       Pittsburgh, PA, USA
 \AND
 \Name{Zhepeng Cen} \Email{zcen@andrew.cmu.edu}
 \\
 \addr Carnegie Mellon University,
       Pittsburgh, PA, USA
\AND
 \Name{Peide Huang} \Email{peideh@andrew.cmu.edu}
 \\
 \addr Carnegie Mellon University,
       Pittsburgh, PA, USA
 \AND
 \Name{Tingnan Zhang} \Email{tingnan@google.com}
 \\
 \addr Google DeepMind,
       Mountain View, CA, USA 
 \AND
 \Name{Wenhao Yu} \Email{magicmelon@google.com}
 \\
 \addr Google DeepMind,
       Mountain View, CA, USA 
 \AND
 \Name{Ding Zhao} \Email{dingzhao@andrew.cmu.edu}
 \\
  \addr Carnegie Mellon University,
       Pittsburgh, PA, USA
}

\begin{document}

\maketitle
\vspace{-35pt}
\begin{abstract}%
Online safe reinforcement learning (RL) involves training a policy that maximizes task efficiency while satisfying constraints via interacting with the environments. In this paper, our focus lies in addressing the complex challenges associated with solving multi-constraint (MC) safe RL problems. We approach the safe RL problem from the perspective of Multi-Objective Optimization (MOO) and propose a unified framework designed for MC safe RL algorithms. This framework highlights the manipulation of gradients derived from constraints. Leveraging insights from this framework and recognizing the significance of \textit{redundant} and \textit{conflicting} constraint conditions, we introduce the Gradient Shaping (GradS) method for general Lagrangian-based safe RL algorithms to improve the training efficiency in terms of both reward and constraint satisfaction. Our extensive experimentation demonstrates the effectiveness of our proposed method in encouraging exploration and learning a policy that improves both safety and reward performance across various challenging MC safe RL tasks as well as good scalability to the number of constraint. 
% The full paper with appendix and video of our work are available on the website:  
% \href{https://sites.google.com/view/mc-grads/home}{\textcolor{blue}{https://sites.google.com/view/mc-grads/home.}
% } # removing this for the arxiv version
\end{abstract}


\vspace{-10pt} 
\section{Introduction}
\vspace{-3pt} 
% Despite the great success of deep reinforcement learning in recent years~\citep{mnih2013playing,silver2017mastering,duan2016benchmarking}, Ensuring safety (i.e., constraint satisfaction) is one key challenge when deploying them to real-world applications~\citep{liu2022constrained}. Safe reinforcement learning (RL) has emerged as a promising approach to address the challenges faced by agents operating in complex, safety-critical environments~\citep{gu2022review}, such as autonomous driving~\citep{isele2018safe}, home service~\citep{ding2022safe}, and UAV locomotion~\citep{qin2021learning}. Safe RL aims to learn a reward-maximizing policy within a constrained policy set~\citep{yang2022safe, thananjeyan2021recovery, bharadhwaj2020conservative, khattar2022cmdp, ding2023provably, wei2023provably}.
% By explicitly accounting for safety constraints during policy learning, agents can better reason about the trade-off between task performance and safety constraints, making them well-suited for real-world applications~\citep{brunke2021safe}.

Despite the great success of deep reinforcement learning (RL) in recent years~\citep{levine2020offline,silver2017mastering,brunke2022safe, li2023deep}, ensuring safety (i.e., constraint satisfaction) is one key challenge when deploying them to real-world applications~\citep{hu2023learning, liu2022constrained, zhao2021model,xu2022trustworthy,wachi2020safe, zhang2023exact}. Safe RL has been a common approach to address the difficulties faced by agents operating in complex and safety-critical tasks~\citep{gu2022review,thananjeyan2021recovery,zhang2020first,zhao2023probabilistic,cheng2023safe,wachi2021safe}, such as autonomous driving~\citep{isele2018safe, hsu2023isaacs}, home service~\citep{ding2022safe, hsu2023sim}, legged robots~\citep{kim2023not}, and UAV locomotion~\citep{qin2021learning, zheng2021safe}. Safe RL aims to maximize the cumulative reward within a constrained policy set~\citep{yang2022safe, thananjeyan2021recovery, bharadhwaj2020conservative, khattar2022cmdp, yao2023constraint, ma2022joint}.
By explicitly incorporating safety constraints into the policy learning process, agents can adeptly navigate the trade-off between task performance and safety constraints, rendering them well-suited for real-world tasks.~\citep{brunke2021safe, yao2023constraint}.

In real-world applications, agents often face multiple constraints \citep{kim2023not, lin2023safetyaware}. For example, an autonomous driving vehicle must avoid collisions, prevent over-speeding, stay on the road, and adhere to various traffic rules and social norms simultaneously~\citep{feng2023dense}. Nevertheless, despite the advancements in safe RL, the development of algorithms for MC safe learning that can effectively handle multiple costs remains a challenging issue~\citep{kim2023trust}. Many existing methods only consider a single constraint during training \citep{achiam2017constrained}. The extension of the Lagrangian method to MC settings is a potential solution. However, such approaches can be sensitive to the initialization of Lagrange multipliers and the learning rate, leading to extensive hyperparameter tuning costs \citep{xu2021crpo, achiam2017constrained, chow2019lyapunov}. Furthermore, these methods may introduce instability issues in scenarios with multiple constraints, thus limiting their scalability. CRPO method~\citep{xu2021crpo} has been proposed to randomly select one constraint for policy consideration at each step to handle multiple constraints. Unfortunately, 
% the efficiency of training is compromised as 
considering one constraint at a time becomes inefficient with an increasing number of constraints.

Empirical findings have indicated that MC safe RL poses more challenges compared to single-cost settings \citep{liu2023datasets, kim2023trust}. In this study, we analyze the MC safe RL problem through the lens of constraint types, identifying two challenging MC safe RL settings: \textit{redundant} and \textit{conflicting} constraints. To address these challenges, we propose the constraint gradient shaping (GradS) technique from the standpoint of Multi-Objective Optimization (MOO), ensuring compatibility with general Lagrangian-based safe RL algorithms. The main contributions are summarized as follows:

% \textbf{1. Identification of challenges induced by MC safe RL settings from the perspective of constraint gradients and their relationship:} We define redundant and conflicting constraints, which lead to over-conservativeness and being stuck in local optima, respectively, thereby hindering online safe RL training. We demonstrate that directly extending methods from single-constraint to multi-constraint tasks is ineffective in these challenging scenarios.

\textbf{1. We introduce a unified framework for Lagrangian-based MC safe RL algorithms} from the perspective of Multi-Objective Optimization (MOO). Within this framework, the major difference among Lagrangian-based MC safe RL methods is the strategy dealing with gradients induced by constraints.

\textbf{2. We propose the gradient shaping (GradS) method for MC safe RL algorithms.} The proposed method can tackle the challenging \textit{redundant} and \textit{conflicting} MC safe RL settings. Our theoretical analysis further provides insights into the convergence of our approach.

\textbf{3. We conduct extensive evaluations of our method:} The proposed GradS method and baselines are evaluated on the MC safe RL tasks modified from common safe RL benchmarks \texttt{Bullte-Safety-Gym} \citep{gronauer2022bullet} and \texttt{Safety-Gymnasium} \citep{Safety-Gymnasium}. The results demonstrate that GradS can significantly improve safety and reward performance in MC tasks.

\vspace{-10pt} 
\section{Related Work}
\vspace{-3pt} 
\textbf{Safe RL} has been approached through various methods. 
% Some techniques leverage domain knowledge of the target problem to enhance the safety of an RL agent~\citep{dalal2018safe, saunders2017trial, alshiekh2018safe, yu2022towards, liu2020safe, luo2021learning, sootla2022saute, chen2021context, pmlr-v202-liu23l}. 
Researchers have proposed many techniques employing constrained optimization techniques to learn a constraint-satisfaction policy~\citep{garcia2015comprehensive, gu2022review, flet2022saac}, such as the Lagrangian-based approach~\citep{bhatnagar2012online, chow2017risk, as2022constrained,ding2023provably}, where the Lagrange multipliers can be optimized along with the policy parameters~\citep{liang2018accelerated, tessler2018reward, ray2019benchmarking}. Alternatively, some works approximate the constrained RL problem with Taylor expansions~\citep{achiam2017constrained} or through variational inference~\citep{liu2022constrained}. They then solve for the dual variable using convex optimization~\citep{yu2019convergent, yang2020projection, gu2021multi, kim2022efficient}. For MC settings, many works propose to consider all the constraints equally \citep{fernando2022mitigating, as2022constrained}, some techniques consider the constraints that violate the most, and other methods randomly activate one constraint for policy update. One recent concurrent work \citep{kim2023trust} proposes the gradient integration method to manage infeasibility issues in MC Safe RL. However, this method is limited to the TRPO-based methods and is hard to generalize to other algorithms. The systematical analysis for MC safe RL is still a largely unexplored area.
% However, most existing approaches consider a fixed constraint threshold during training, which can hardly be deployed for different threshold conditions after training.
\\
\textbf{Multi-Objective Optimization (MOO)} considers how to train a single model that can meet a variety of different requirements \citep{huang2022constrained, yu2020gradient, caruana1997multitask}. The MOO formulation has been extended to many different settings, including supervised learning \citep{yang2016trace, zamir2018taskonomy}, and reinforcement learning \citep{wilson2007multi, sodhani2021multi}. For the Multi-Objective RL (MORL), existing works learn a policy that is optimal in the Pareto Frontier with a given trading-off among tasks \citep{roijers2013survey, zhang2020random}. In recent years, researchers also interpreted safe RL from the perspective of MORL. However, they are primarily focusing on multiple task rewards and preference settings \citep{huang2022constrained} and single-constraint settings \citep{liu2023constrained}, but not particular MC safe RL problems.  \\

\vspace{-20pt}
\section{Unified Framework for MC Safe RL}
\vspace{-3pt} 
In this section, we introduce the proposed unified framework for Lagrangian-based MC safe RL.
\vspace{-10pt}
\subsection{Preliminary}
\label{subsection: MD safe RL}
Constrained Markov Decision Process (CMDP) $\Mcal$ is defined by the tuple $(\Scal, \Acal, \Pcal, r, \boldsymbol{c}, \mu_0)$ \citep{altman1998constrained}, where $\Scal$ is the state space, $\Acal$ is the action space, $\Pcal:\Scal \times \Acal \times \Scal \xrightarrow{} [0, 1]$ 
is the transition function, $r:\Scal \times \Acal \times \Scal \xrightarrow{} \mathbb{R}$ is the reward function, and $\mu_0: \Scal \xrightarrow[]{} [0,1]$ is the initial state distribution.
CMDP augments MDP with an additional element $\textbf{c}:\Scal \times \Acal \times \Scal \xrightarrow{} \mathbb{R}^{N}_{\geq 0}
$ to characterize the cost of violating the constraint, where $N$ is the cost dimension. An MC safe RL problem is specified by a CMDP and a constraint threshold vector $\boldsymbol{\epsilon} \in \mathbb{R}^{N}_{\geq 0}$.
Let $\pi:\Scal \times \Acal\rightarrow [0,1]$ denote the policy and $\tau = \{s_1, a_1, ...\}$ denote the trajectory.
% We use shorthand $\fs_t=\fs(s_t, a_t, s_{t+1}), \fs\in\{r, \boldsymbol{c}\}$ for simplicity. 
The value functions are $V_r^\pi(\mu_0) = \mathbb{E}_{\tau \sim \pi, s_0 \sim \mu_0}[ \sum_{t=0}^\infty \gamma^t r(t) ], \boldsymbol{V}_{c_i}^\pi(\mu_0) = \mathbb{E}_{\tau \sim \pi, s_0 \sim \mu_0}[ \sum_{t=0}^\infty \gamma^t \boldsymbol{c}_i(t) ], i=1, 2, ..., N$, which is the expectation of discounted return under the policy $\pi$ and the initial state distribution $\mu_0$. Denote $\preceq$ as an element-wise partial order, the goal of MC safe RL is to find the policy that maximizes the reward return while constraining the cost return under the pre-defined threshold $\boldsymbol{\epsilon}$:
\begin{wrapfigure}{R}{0.42\textwidth}
% \vspace*{-0.10in}
\centering
\includegraphics[width=.98\linewidth]{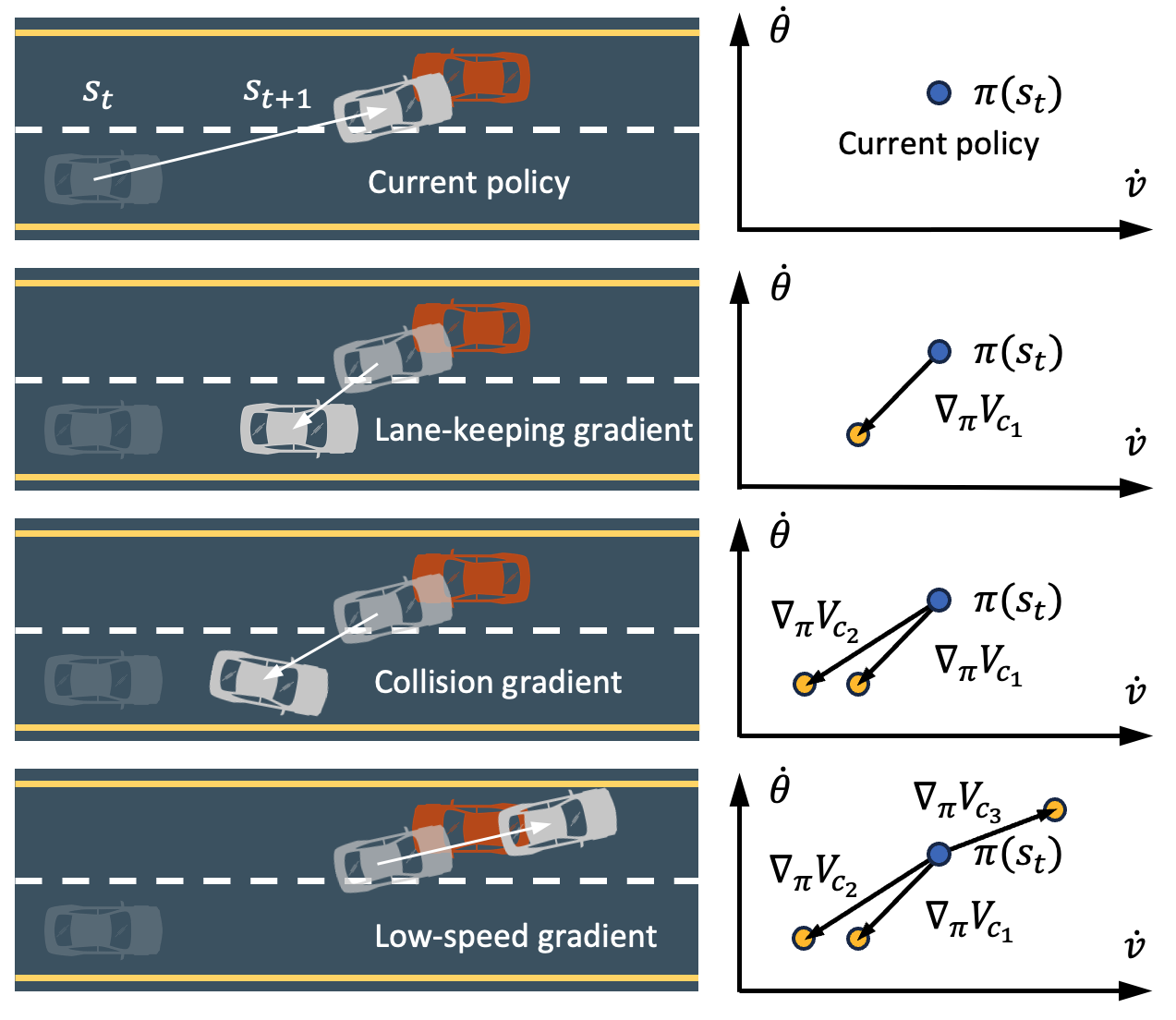}
\caption{ Illustration of constraint types. $c_1$ is the lane-keeping cost, $c_2$ is the collision avoidance cost, and $c_3$ is the low-speed cost. 
% $c_1$ and $c_2$ are redundant, $c_1$ and $c_3$ are conflicting.
}
\label{Fig: constraint type}
% \vspace*{-0.05in}
\end{wrapfigure}
\begin{equation}
% \vspace{-2mm}
\label{equ: MD safe RL def}
   \pi^* = \arg\max_{\pi}  V_r^\pi, \quad s.t. \quad \boldsymbol{V}_{\boldsymbol{c}}^\pi \preceq \boldsymbol{\epsilon}, \quad (\boldsymbol{V}_{\boldsymbol{c}}^\pi \in \mathbb{R}^{N}_{\geq 0}, \ \boldsymbol{\epsilon} \in \mathbb{R}^{N}_{\geq 0})
\end{equation} 
To solve this problem, Lagrangian-based safe RL algorithms can be formulated to find:
\begin{equation}
\left(\mathbf{\pi}^*, \boldsymbol{\lambda}^*\right)=\arg \max _\lambda \min _{\mathbf{\pi}} \mathcal{J}(\mathbf{\pi}, \boldsymbol{\lambda}), \quad \mathcal{J}(\mathbf{\pi}, \boldsymbol{\lambda}) = - V_r^\pi + \boldsymbol{\lambda}^T \boldsymbol{V}_{\boldsymbol{c}}^\pi
\label{equ: lagrangian safe RL}
\end{equation}
where $\boldsymbol{\lambda} = [\lambda_1, \lambda_2, ..., \lambda_N]^T$ is the Lagrangian multiplier corresponding to the primary problem (\ref{equ: MD safe RL def}). In practice, we can update $\left(\mathbf{\pi}, \boldsymbol{\lambda}\right)$ iteratively \citep{stooke2020responsive}.

\subsection{Unified framework: MC Safe RL as MOO}
\label{subsection: MC Safe RL as MOO}
In multi-objective optimization (MOO), we are given $K \geq 2$ different tasks, each associated with a loss function \citep{fernando2022mitigating}. With this, at $t$-th step, updating $\pi_t$ via solving~(\ref{equ: lagrangian safe RL}) is to find:
\begin{equation}
    \pi_t^* = \arg\min_{\pi_t} \left[ -V_r^{\pi_t} + \boldsymbol{\lambda_t}^T \boldsymbol{V}_{\boldsymbol{c}}^{\pi_t} \right],
\end{equation}
For simplicity, we will omit the subscript $t$ and superscript $\pi$ in the following. The gradient $\nabla J$ for policy $\pi$ is:
% \begin{equation}
% \label{equ: unshaped gradient framework}
%     \nabla J = -\nabla_{\pi} V_r^{\pi} + \boldsymbol{\lambda}^T \nabla_{\pi} \boldsymbol{V}_{\boldsymbol{c}}^\pi
% \end{equation}
% Then reformulate the gradient in~(\ref{equ: unshaped gradient framework}) as:
\begin{equation}
\label{equ: unshaped gradient framework MOO}
    \nabla J = -\nabla V_r + \nabla J_c, \quad \nabla J_c = \boldsymbol{w}^T \boldsymbol{G},
\end{equation}
where $\boldsymbol{G} := [ g_1, ..., g_m]$ is the constraint gradient vector, $g_i = \boldsymbol{\lambda}_i \nabla \boldsymbol{V}_{\boldsymbol{c}_i}$ is the $i$-th constraint gradient,  and $\boldsymbol{w} \succeq \boldsymbol{0}$ is a non-negative weight vector of the constraint gradients. With this formulation, many commonly used methods for MC Safe RL can be categorized as:
% $\boldsymbol{G} := [ \boldsymbol{\lambda}_1 \nabla_{\pi} \boldsymbol{V}_{\boldsymbol{c}_1}, ..., \boldsymbol{\lambda}_m \nabla_{\pi} \boldsymbol{V}_{\boldsymbol{c}_m}]$

\textbf{(1) Vanilla Method: }For common safe RL algorithms \citep{fernando2022mitigating, as2022constrained}, they consider all the constraints equally, with a uniform weight:
\begin{equation}
    \boldsymbol{w} = \boldsymbol{1}
\end{equation}

\textbf{(2) CRPO\footnote{We modify the original CRPO to a Lagrangian version. Please refer to the experiment and appendix for more details.} Method: }
% Denote $\in_R$ as the random sampling operation, then 
Methods such as CRPO \citep{xu2021crpo} that randomly select constraints for policy update at each time can be formulated as:
\begin{equation}
    \| \boldsymbol{w} \|_0 = 1, \quad  \ \boldsymbol{w}_i = 1,\ i \sim \text{uniform}(1, N)
\end{equation}

\textbf{(3) Min-Max method: }Safe RL methods that penalize the cost  that violates the constraint the most for policy updates at each time can be formulated as:
\begin{equation}
    \| \boldsymbol{w} \|_0 = 1, \quad  \ \boldsymbol{w}_i = 1,\ i = \arg\max (\boldsymbol{V}_{{c}_i} - \boldsymbol{\epsilon}_i)
\end{equation}

\vspace{-10pt} 
\section{Gradient Shaping for MC Safe RL}
\vspace{-3pt} 
Based on empirical findings in both previous works \citep{liu2023datasets, kim2023trust} and this work, MC safe RL presents greater difficulty compared to single-constraint ones. Thus, before delving into the proposed method, we outline the critical conditions essential for understanding MC safe RL, particularly focusing on various constraint types.

% \begin{wrapfigure}{R}{0.55\textwidth}
% \begin{minipage}{0.54\textwidth}
% % \begin{figure}
%     % \centering
%     \includegraphics{Figures/Illustration/Toy-example.png}
%     \caption{Caption}
%     \label{fig:enter-label}
% % \end{figure}
% \end{minipage}
% \end{wrapfigure}

\subsection{Constraint Types in MC Safe RL }

\begin{wrapfigure}{R}{0.45\textwidth}
\vspace*{-0.40in}
\centering
\includegraphics[width=.98\linewidth]{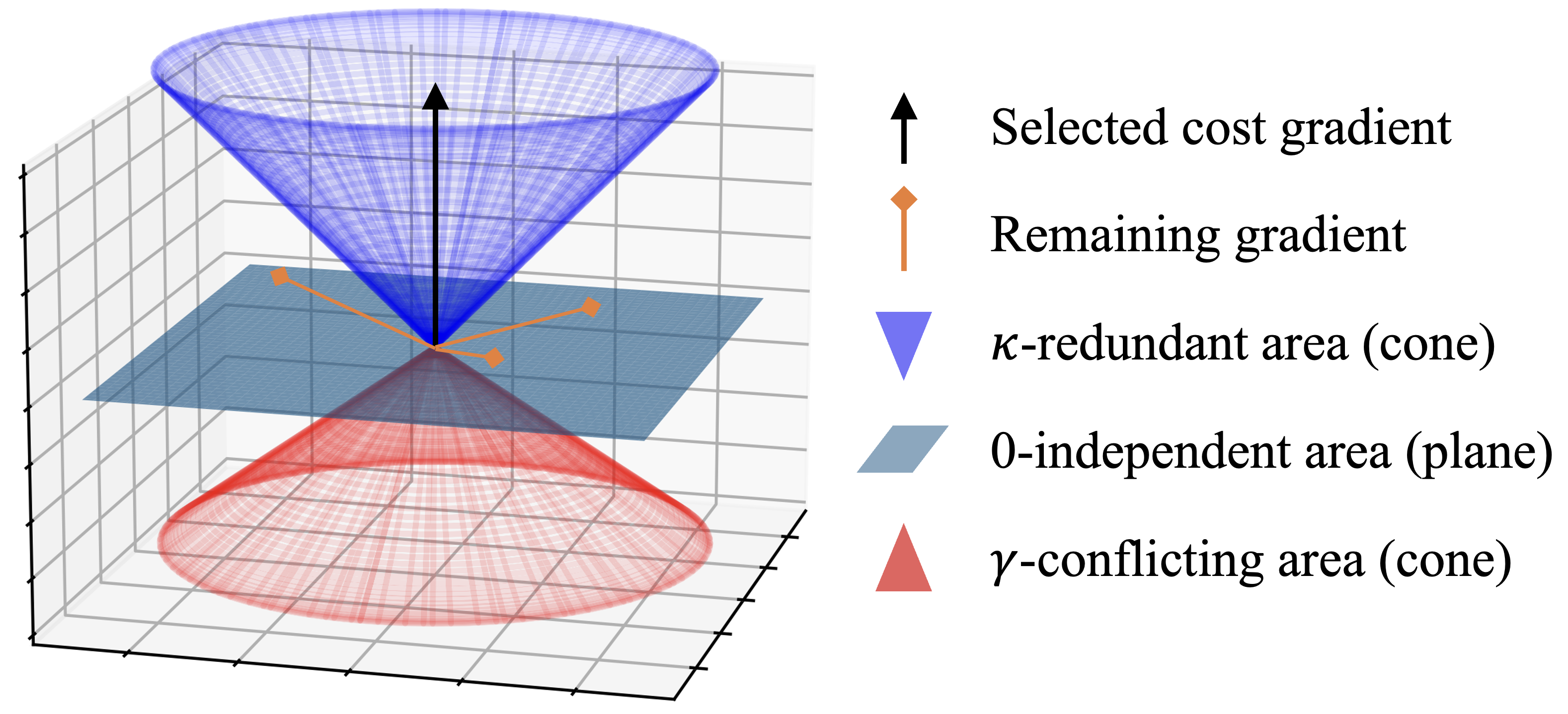}
\caption{Illustration of elimination area.}
\label{Fig: constraint area}
\vspace*{-0.25in}
\end{wrapfigure}

Based on the constraint gradient similarity, we define the relationship between two distinct constraints $\boldsymbol{V}_{\boldsymbol{c}_i}^\pi \leq \boldsymbol{\epsilon}_i$ and $\boldsymbol{V}_{\boldsymbol{c}_j}^\pi \leq \boldsymbol{\epsilon}_j$ for $i \neq j$ given a policy $\pi$. Note that the gradients are closely related to the current policy $\pi$. We utilize the cosine similarity, which has been used in many previous works \citep{du2018adapting}, as the similarity function $sim(\cdot, \cdot)$. Denote $\theta$ as the parameter for the policy $\pi$.

\begin{wrapfigure}{R}{0.50\textwidth}
\begin{minipage}{0.50\textwidth}
\vspace*{-0.30in}
\begin{algorithm}[H]
\caption{Gradient Shaping (GradS)}
{\bfseries Input:} \raggedright policy $\pi$ \par
{\bfseries Output:} \raggedright shaped constraint gradient $\nabla J_c$  \par
\begin{algorithmic}[1] % The number tells where the line numbering should start
% \STATE $\triangleright$ \textit{compute cost sampling range}
% \STATE $\triangleright$ \textit{\textcolor{blue}{Gradient shaping begins}}
% \STATE $\triangleright$ \textit{Shuffling the constraint index:}
\STATE Shuffling the constraint indices
\STATE $\triangleright$ \textit{Initialize the candidate gradient set} 
\STATE $\mathcal{G}\leftarrow \{g_1:=\lambda_1 \nabla V_{c_1} \}$
\STATE $\triangleright$ \textit{Get the candidate gradient set }$\mathcal{G}$
\FOR{$i=2,..., n$}
\IF {$-\sigma < sim(i, j) < \kappa, \ \forall j \in \{ \mathcal{G}\}$}
\STATE $\triangleright$ \textit{Add this constraint into the set}
\STATE $\mathcal{G}\leftarrow \mathcal{G} \cup \{g_i:=\lambda_i \nabla V_{c_i} \}$
\ENDIF
\ENDFOR
\STATE $\triangleright$ \textit{Select constraint gradient}
% \STATE $k \in_{R} \text{index}(\lambda_\mathcal{G})$
\STATE $g_c \sim \text{uniform}(\mathcal{G})$
% \STATE $\triangleright$ \textit{\textcolor{blue}{Gradient shaping ends}}
\STATE {\bfseries Return:} $\nabla J_c = \nabla V_c^{G} = g_c \ |\mathcal{G}| / N$ 
\end{algorithmic} \label{algo: gradient shaping}
\end{algorithm}
\end{minipage}
\vspace*{-0.25in}

\end{wrapfigure}

\begin{definition}[\textbf{$\sigma$-conflicting constraints}] The constraints $\boldsymbol{V}_{\boldsymbol{c}_i}^\pi \leq \boldsymbol{\epsilon}_i$ and $\boldsymbol{V}_{\boldsymbol{c}_j}^\pi \leq \boldsymbol{\epsilon}_j$ are $\sigma$-conflicting constraints if and only if:
\begin{equation}
    {sim}(\nabla_{\theta} \boldsymbol{V}_{\boldsymbol{c}_i}^\pi, \nabla_{\theta} \boldsymbol{V}_{\boldsymbol{c}_j}^\pi) \leq -\sigma,
\end{equation}
Conflicting constraints drive the policy in conflicting directions if both are activated.
    
\end{definition}

\begin{definition}[\textbf{$\kappa$-redundant constraints}] The constraints $\boldsymbol{V}_{\boldsymbol{c}_i}^\pi  \leq \boldsymbol{\epsilon}_i$ and $\boldsymbol{V}_{\boldsymbol{c}_j}^\pi  \leq \boldsymbol{\epsilon}_j$ are $\kappa$-redundant constraints if and only if:
\begin{equation}
    {sim}(\nabla_{\theta} \boldsymbol{V}_{\boldsymbol{c}_i}^\pi , \nabla_{\theta} \boldsymbol{V}_{\boldsymbol{c}_j}^\pi ) \geq \kappa,
\end{equation}
Redundant constraints drive the policy in almost the same direction if both are activated. 
\end{definition}

\begin{definition}[\textbf{$\eta$-independent constraints}] The constraints $\boldsymbol{V}_{\boldsymbol{c}_i}^\pi  \leq \boldsymbol{\epsilon}_i$ and $\boldsymbol{V}_{\boldsymbol{c}_j}^\pi  \leq \boldsymbol{\epsilon}_j$ are $\eta$-independent constraint if and only if:
\begin{equation}
     -\eta \leq {sim}(\nabla_{\theta} \boldsymbol{V}_{\boldsymbol{c}_i}^\pi , \nabla_{\theta} \boldsymbol{V}_{\boldsymbol{c}_j}^\pi ) \leq \eta,
\end{equation}
Independent constraints drive the policy in ``independent'' directions if both are activated. 
\end{definition}
% Note that redundant constraints harm multi-constraint settings by inducing over-conservativeness in the training.

% We should note that the gradient similarity can be defined in certain feature spaces, including the agent action space and the policy space (Neural network parameter space). 
% In practice, we calculate the similarity in the policy actor parameter space.
For simplicity, we will omit the subscript $\theta$ in the following context.
With the toy example in Figure. \ref{Fig: constraint type}, we illustrate the aforementioned \textit{redundant} and \textit{conflicting} constraints, which are two primary optimization issues in MC safe RL. In this common autonomous driving scenario, we consider three constraints: the lane-keeping constraint to keep the car on the lane, the collision constraint to prevent accidents with other vehicles, and the minimum speed limit constraint to prevent congestion. In the case shown in Figure. \ref{Fig: constraint type}, for current policy, the lane-keeping constraint $c_1$ and the collision constraint $c_2$ are redundant, while $c_1$ and the low-speed constraint $c_3$ are conflicting. 
Notably, \textit{redundant} and \textit{conflicting} constraints are not inherently problematic. In fact, simply averaging constraint gradients should lead to the optimal policy for MC safe RL problems. 
However, for online safe RL algorithms, \textit{redundant} constraints lead to over-conservativeness by over-estimating the effect of constraints, while \textit{conflicting} constraints result in exploration instability as getting stuck in local optimum, both of which are detrimental to online safe RL agent learning.

\subsection{Gradient Shaping}
\begin{figure}[t]
    \centering
    \vspace{-8pt} 
    \includegraphics[width=0.99\linewidth]{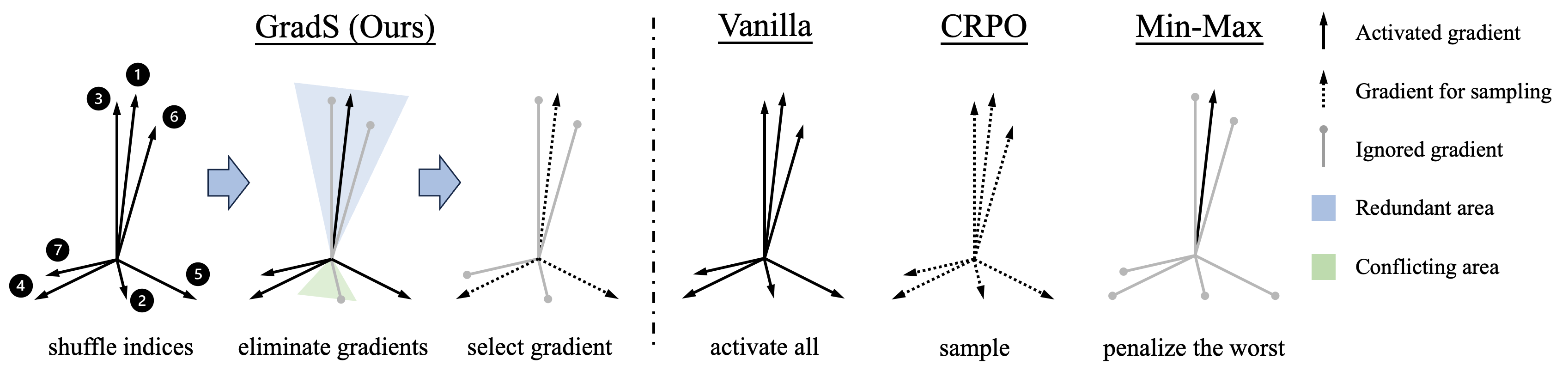}
    \caption{\small Illustration for constraint gradients shaping.}
    \label{fig:GradS_illustration}
\end{figure}

The objective of our approach is to address the challenges posed by \textit{redundant} and \textit{conflicting} constraints, aiming to eliminate over-conservativeness resulting from \textit{redundant} constraints and escape local optima to resolve \textit{conflicting} constraints. In this section, we outline our strategy for shaping the constraint gradients. We also provide a theoretical analysis demonstrating that GradS still guarantees convergence in the next section. The core idea for GradS is to first get a candidate constraint gradient set $\mathcal{G}$ via eliminating the \textit{redundant} and \textit{conflicting} constraints, then randomly select one constraint gradient in set $\mathcal{G}$ for policy update. The proposed algorithm operates as follows:

(1) Initially, it shuffles the constraint gradients and computes the cosine similarity between each pair of constraint gradients $\nabla V_{c_i}$ and $\nabla V_{c_j}$.
% to identify whether they are $\kappa$-redundant constraints.
(2) Next, it initializes the candidate gradient set with the first gradient $\mathcal{G} \leftarrow \{ g_1:=\lambda_1 \nabla V_{c_1}\}$.
(3) It then selects gradients sequentially: if a newly chosen gradient $g_i$ is neither $\kappa$-redundant nor $\gamma$-conflicting with any other gradient in the set $\mathcal{G}$, it is added to the set $\mathcal{G}\leftarrow \mathcal{G} \cup \{g_i:=\lambda_i \nabla V_{c_i}\}$. Otherwise, it skips this constraint.
(4) After the selection process, the constraint candidate set $\mathcal{G}$ is obtained. Then it randomly samples a gradient from $\mathcal{G}$, and multiplied by a scaling factor as the constraint gradient $V_{c}^G = g_{\tilde{i}} \ |\mathcal{G}| / N $, where $\tilde{i}$ denotes the index for the selected constraint $\tilde{i} \sim \text{uniform}(\mathcal{G}), g_{\tilde{i}} = \mathcal{G}[\tilde{i}]$. The scaling term $|\mathcal{G}| / N$ is used to ensure stability. The process is described in Algorithm~\ref{algo: gradient shaping}. The illustration of the proposed GradS method and the comparison with baseline methods are shown in Figure \ref{fig:GradS_illustration}.

The GradS method, although straightforward, mitigates constraints by excluding \textit{redundant} and \textit{conflicting} gradients, which induce over-conservativeness and exploration issues for online safe RL, and selects cost gradients that are \textit{independent}, which makes the policy update more efficient as well as considering most cost information as shown in Figure. \ref{Fig: constraint area}. Moreover, it encourages exploration by sampling from the gradients after the elimination process instead of aggregating them. In practice, the GradS method can be applied to general Lagrangian-based safe RL algorithms (discussed in this paper) and has the potential for extension to general safe RL algorithms. The proposed GradS method also falls into the framework (\ref{equ: unshaped gradient framework MOO}) as to find the weight $\boldsymbol{w}$:
\begin{equation}
    \| \boldsymbol{w} \|_0 = 1, \quad \boldsymbol{w}_i = |\mathcal{G}| / N, \quad i \sim \text{uniform}(\text{index}(\mathcal{G})), 
\end{equation}
where $\mathcal{G}$ is the set for candidate cost gradients as mentioned above and shown in Alg. \ref{algo: gradient shaping}, and the sampling ``$\sim$'' means to sample from the corresponding indices of gradients in the candidate set.

\vspace{-8pt} 
\subsection{Theoritical analysis}
In this section, we theoretically analyze the performance of GradS with the convergence guarantee. We first have these two common assumptions in safe RL:
\begin{Assumption}[Slater's condition] The feasible policy exists, i.e., $\exists \pi$, such that $\boldsymbol{V}_{\boldsymbol{c}}^\pi  \preceq \boldsymbol{\epsilon}$.
\end{Assumption}
The feasibility assumption ensures that the Lagrangian $\boldsymbol{\lambda}$ corresponding to the optimization problem (\ref{equ: lagrangian safe RL}) is bounded.

\begin{Assumption}[Bounded and smooth gradients] 
\label{assumption: Smooth and bounded gradient components} Assuming the constraint gradient components are bounded and smooth, i.e., for some constants $G, L > 0$,
\begin{equation}
\left\|\nabla V_{c_i}\right\| \leq G, \quad \left|\boldsymbol{u}^T \nabla^2 V_{c_i} \boldsymbol{u}\right| \leq L\|\boldsymbol{u}\|^2, \ \forall \boldsymbol{u} \in \mathbb{R}^d
\end{equation}
\end{Assumption}
where $\mathbb{R}^d$ characterizes the policy gradient space. With these mild assumptions, we can ensure that the constraint gradient after GradS is still bounded as shown in Theorem \ref{theorem: convergence analysis}.

\begin{theorem}[Convergence analysis] Denote the number of removed $\kappa$-redundant and $\sigma$-conflicting constraints at iteration time step $t$ as $N_R(\kappa, t), N_C(\sigma, t)$, the total optimized time step as $T$, the learning rate for every optimization step is $\alpha$, then for the safety performance, i.e., if we only consider constraint gradient $\nabla V_c^G\left(\theta_t\right)$, the policy gradient can be bounded as
\begin{equation}
\begin{aligned}
    \mathbb{E}_t\left[\left\|\nabla V_c^G\left(\theta_t\right)\right\|^2\right] \leq \frac{V_c\left(\theta_0\right)-V_c^*}{T \alpha} + G^2 \left( \mathbb{E}_t \left[N_R(\kappa, t)\right]+ \mathbb{E}_t \left[N_C(\sigma, t)\right] \right)  +\frac{\alpha G^2 L}{2} 
\end{aligned}
\label{equ: convergence analysis}
\vspace{-20pt}
\end{equation}
\label{theorem: convergence analysis}
\end{theorem}
% The proof can be seen in the appendix of [link]. This bound is comprised of three terms. The first term is related to the initialization parameters; the second term is the result of eliminating the redundant constraints in the gradient shaping; and the third term is caused by randomly selecting gradients in the shaped gradients set. The last two terms are caused by the proposed gradient shaping process. From equation \ref{theorem: convergence analysis}, we can find that the gradient shaping process indeed encourages exploring with constraint gradients by removing the redundant constraints and conflicting constraints, thus benefits the learning process of Online safe RL in the MC settings.
The proof is available in the appendix. This bound consists of three terms. The first term relates to the initialization parameters, the second term arises from the elimination of \textit{redundant} and \textit{conflicting} constraints, and the third term is due to the sampling of gradients from the candidate set. The last two terms result from the proposed GradS, which are our ``noise ball'' terms: the terms that are in some sense ``causing'' GradS to converge not to a point with zero gradients but rather to some reason nearby, thus we can improve the learning efficiency by avoiding getting stuck in local optimum. 
% As evident from equation (\ref{equ: convergence analysis}),  the gradient shaping process indeed promotes exploration with constraint gradients, thereby enhancing the learning efficiency of online safe RL agents in the MC settings.

\vspace{-10pt} 
\section{Experiments}
\vspace{-3pt} 

We aim to address three primary questions in the experiment section: (a) Can baseline methods effectively learn policies that are both safe and rewarding in the challenging MC tasks? (b) How does the proposed GradS method perform in the MC environments? (c) What is the scalability of the proposed GradS method concerning the number of constraints in safe RL tasks? To answer these questions, we employ the following experiment setup to assess GradS and the baseline approaches.

\vspace{-8pt} 
\subsection{Experiment setup}
\label{subsection: Experiment setup}

\begin{wrapfigure}{R}{0.4\textwidth}
\vspace*{-0.05in}
\centering
\includegraphics[width=.98\linewidth]{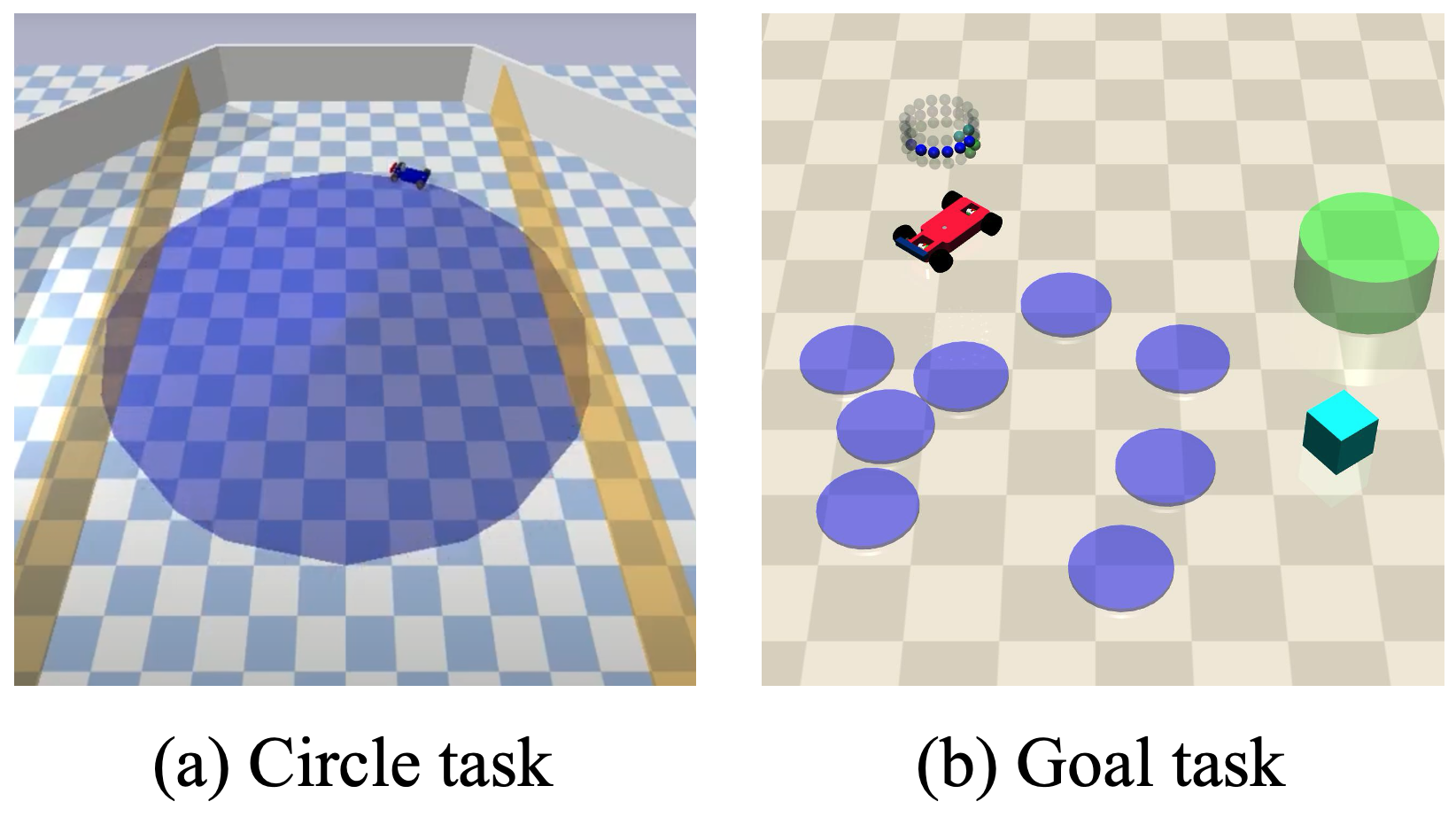}
\vspace*{-0.05in}
\caption{Task visualization.}
\label{Fig: tasks}
\vspace*{-0.1in}
\end{wrapfigure}

\textbf{Tasks.} We utilize several continuous control tasks for robot locomotion commonly employed in previous studies \citep{achiam2017constrained, chow2019lyapunov, zhang2020first}. The simulation environments are sourced from public benchmark \texttt{Bullet-Safety-Gym} \citep{gronauer2022bullet} and \texttt{Safety-Gymnasium} \citep{Safety-Gymnasium}. We consider two tasks (\texttt{Circle} and \texttt{Goal}) as shown in Figure \ref{Fig: tasks} and train with various robots (\texttt{Point, Ball, Car}, and \texttt{Drone}). In the \texttt{Circle} environment, agents are rewarded for following a circular path. In the \texttt{Goal} task, agents are rewarded for reaching the goal cube. The details of the tasks can be found in the appendix. We name the task as ``Robot''-``task'', for example, \texttt{BC} means ``\underline{B}all-\underline{C}ircle'', and \texttt{CG} means ``\underline{C}ar-\underline{G}oal''.

\textbf{Constraints.} In the aforementioned tasks, the original environment only provides single-dimensional cost information. To better simulate real-world scenarios, we introduce three representative costs: \textbf{Boundary/collision cost:} agents incur a cost if they cross the boundary or collide with the obstacles. \textbf{High-velocity cost:} agents receive a cost if they exceed the upper-velocity limit. \textbf{Low-velocity cost:} agents receive a cost if their speed falls below the lower-velocity limit. All costs are binary. A detailed explanation of the costs is provided in the appendix. Intuitively, boundary/collision cost and high-velocity cost are likely \textit{redundant} constraints since high speed might also result in crossing the boundary or collision. High-velocity cost and low-velocity cost are likely \textit{conflicting} constraints as they potentially tend to pull the policy in conflicting optimization directions if both are activated. 
We create tasks considering the first two types of constraints with the suffix ``\texttt{-v2}'', and tasks with all three types of constraints with the suffix ``\texttt{-v3}'' (more challenging). 
% The scalability experiment results shown in Figure. \ref{fig: Scalability analysis}, we increase the dimension of constraint by creating new constraints with slightly different velocity upper bounds. More details about the experiments can be found in the Appendix.

\textbf{Metrics. }We compare the methods in terms of episodic reward (the higher, the better) and episodic constraint cost violations (the lower, the better), which have been used in many related works~\citep{liu2023constrained, li2023guided}. We normalized the cost, and reported the most-violated cost among all the constraints (then the cost threshold becomes $1$): 
\begin{equation}
   \text{cost-N} = \max_i \{ c_i / \epsilon_i \}
\end{equation}

\textbf{Algorithms and baselines. }For the safe RL algorithms, we select commonly used model-free off-policy algorithms, \texttt{SAC-Lag} and \texttt{DDPG-Lag}, and model-free on-policy methods, \texttt{PPO-Lag} and \texttt{TRPO-Lag}. As introduced in Section \ref{subsection: MC Safe RL as MOO}, the baseline methods are \texttt{Vanilla}, \texttt{CRPO}, and \texttt{Min-Max}. For the CRPO method, we modify it to a Lagrangian version for a fair comparison. More details and results including practical implementation, and training curves are provided in the appendix.

% % \tiny
% {
% \begin{table}[htbp]
%   \centering
%   \caption{ \small Add caption}
%     \begin{tabular}{rccccccccc}
%     \toprule
%     \multicolumn{1}{c}{\multirow{2}[4]{*}{Env}} & \multirow{2}[4]{*}{Method} & \multicolumn{2}{c}{GradS (ours)} & \multicolumn{2}{c}{Vanilla} & \multicolumn{2}{c}{Sampling} & \multicolumn{2}{c}{Min-Max} \\
% \cmidrule{3-10}          &       & Reward & Cost-N & Reward & Cost-N & Reward & Cost-N & Reward & Cost-N \\
%     \midrule
%           & PPO-L & 271.63 ± 24.08 & 1.03 ± 0.12 & 260.2 ± 19.44 & 0.88 ± 0.17 & 288.04 ± 15.79 & 1.85 ± 0.11 & 245.41 ± 21.18 & 1.16 ± 0.12 \\
%     \multicolumn{1}{c}{Ball-v3ircle} & TRPO-L & 329.85 ± 10.39 & 1.20 ± 0.06 & 283.88 ± 35.16 & 1.01 ± 0.04 & 362.35 ± 5.87 & 4.51 ± 0.11 & 331.41 ± 10.69 & 1.24 ± 0.11 \\
%     \multicolumn{1}{c}{Redundant} & SAC-L & 229.00 ± 33.64 & 1.09 ± 0.24 & 228.83 ± 16.61 & 1.18 ± 0.29 & 271.92 ± 28.16 & 1.49 ± 0.26 & 223.45 ± 7.20 & 1.13 ± 0.24 \\
%           & DDPG-L & 170.10 ± 50.92 & 1.06 ± 0.18 & 177.46 ± 48.51 & 1.03 ± 0.15 & 195.66 ± 32.35 & 0.97 ± 0.33 & 202.16 ± 29.17 & 1.11 ± 0.19 \\
% \cmidrule{2-10}          & Average & 250.15 & 1.1   & 237.59 & 1.03  & 279.49 & 2.38  & 200.61 & 1.16 \\
%     \bottomrule
%     \end{tabular}%
%   \label{tab:addlabel}%
% \end{table}%
% }

\begin{table}[ht]
\vspace{-8pt} 
\centering
\renewcommand{\arraystretch}{1.1}
\resizebox{1.\linewidth}{!}{
\begin{tabular}{rccccccccc}
    \toprule
    \multicolumn{1}{c}{\multirow{2}[4]{*}{Env}} & \multirow{2}[4]{*}{Method} & \multicolumn{2}{c}{GradS (ours)} & \multicolumn{2}{c}{Vanilla} & \multicolumn{2}{c}{CRPO} & \multicolumn{2}{c}{Min-Max} \\
\cmidrule{3-10}          &       & Reward $\uparrow$ & Cost-N $\downarrow$ & Reward $\uparrow$ & Cost-N $\downarrow$ & Reward $\uparrow$ & Cost-N $\downarrow$ & Reward $\uparrow$ & Cost-N $\downarrow$ \\
    \midrule
          & PPO-L & 271.63 ± 24.08 & 1.03 ± 0.12 & 260.2 ± 19.44 & 0.88 ± 0.17 & 288.04 ± 15.79 & 1.85 ± 0.11 & 245.41 ± 21.18 & 1.16 ± 0.12 \\
          & TRPO-L & 329.85 ± 10.39 & 1.20 ± 0.06 & 283.88 ± 35.16 & 1.01 ± 0.04 & 362.35 ± 5.87 & 4.51 ± 0.11 & 331.41 ± 10.69 & 1.24 ± 0.11 \\
    \multicolumn{1}{c}{BC-v2} & SAC-L & 229.00 ± 33.64 & 1.09 ± 0.24 & 228.83 ± 16.61 & 1.18 ± 0.29 & 271.92 ± 28.16 & 1.49 ± 0.26 & 223.45 ± 7.20 & 1.13 ± 0.24 \\
          & DDPG-L & 170.10 ± 50.92 & 1.06 ± 0.18 & 177.46 ± 48.51 & 1.03 ± 0.15 & 195.66 ± 32.35 & 0.97 ± 0.33 & 202.16 ± 29.17 & 1.11 ± 0.19 \\
\cmidrule{2-10}          & Average & \cellcolor[rgb]{ .851,  .851,  .851}\textbf{250.15} & \cellcolor[rgb]{ .851,  .851,  .851}\textbf{1.10} & \textbf{237.59} & \textbf{1.03} & \cellcolor[rgb]{ .851,  .851,  .851}279.49 & \cellcolor[rgb]{ .851,  .851,  .851}2.38 & 200.61 & 1.16 \\
\midrule
          & PPO-L & 220.71 ± 9.63 & 1.04 ± 0.15 & 137.02 ± 15.09 & 0.61 ± 0.26 & 233.20 ± 10.11 & 1.85 ± 0.16 & 165.67 ± 29.97 & 0.81 ± 0.24 \\
          & TRPO-L & 242.69 ± 8.69 & 1.01 ± 0.06 & 218.47 ± 13.85 & 1.03 ± 0.11 & 266.01 ± 7.91 & 2.14 ± 0.16 & 239.51 ± 9.72 & 1.01 ± 0.05 \\
    \multicolumn{1}{c}{CC-v2} & SAC-L & 175.69 ± 104.28 & 1.40 ± 0.86 & 34.91 ± 62.89 & 5.57 ± 12.24 & 146.39 ± 57.96 & 1.12 ± 0.67 & 46.39 ± 57.96 & 1.08 ± 1.35 \\
          & DDPG-L & 227.89 ± 1.32 & 1.00 ± 0.32 & 176.62 ± 17.85 & 1.00 ± 0.14 & 237.92 ± 8.57 & 1.93 ± 0.09 & 175.75 ± 18.65 & 1.13 ± 0.69 \\
\cmidrule{2-10}          & Average & \cellcolor[rgb]{ .851,  .851,  .851}\textbf{216.75} & \cellcolor[rgb]{ .851,  .851,  .851}\textbf{1.11} & 147.76 & 2.05  & \cellcolor[rgb]{ .851,  .851,  .851}218.38 & \cellcolor[rgb]{ .851,  .851,  .851}1.76 & \textbf{156.83} & \textbf{1.01} \\
\midrule
          & PPO-L & 253.21 ± 65.49 & 0.88 ± 0.15 & 137.87 ± 36.57 & 0.80 ± 0.16 & 186.30 ± 59.54 & 0.98 ± 0.26 & 164.19 ± 44.54 & 0.98 ± 0.13 \\
          & TRPO-L & 404.16 ± 41.15 & 0.93 ± 0.14 & 306.14 ± 67.78 & 0.89 ± 0.09 & 414.74 ± 69.34 & 1.72 ± 0.83 & 359.51 ± 58.89 & 0.84 ± 0.09 \\
    \multicolumn{1}{c}{DC-v2} & SAC-L & 413.30 ± 76.61 & 0.96 ± 0.12 & 281.47 ± 76.09 & 0.71 ± 0.45 & 544.76 ± 68.24 & 3.01 ± 0.28 & 211.14 ± 58.74 & 0.70 ± 0.22 \\
          & DDPG-L & 399.05 ± 44.12 & 0.92 ± 0.12 & 195.77 ± 44.53 & 0.94 ± 0.14 & 555.65 ± 52.31 & 3.12 ± 0.23 & 234.45 ± 16.91 & 0.84 ± 0.14 \\
\cmidrule{2-10}          & Average & \cellcolor[rgb]{ .851,  .851,  .851}\textbf{367.43} & \cellcolor[rgb]{ .851,  .851,  .851}\textbf{0.92} & \textbf{230.3} & \textbf{0.84} & \cellcolor[rgb]{ .851,  .851,  .851}425.36 & \cellcolor[rgb]{ .851,  .851,  .851}2.21 & \textbf{242.32} & \textbf{0.84} \\
    \midrule
          & PPO-L & 214.00 ± 57.16 & 0.98 ± 0.12 & 40.52 ± 25.33 & 1.02 ± 0.54 & 339.23 ± 72.88 & 1.85 ± 0.49 & 28.89 ± 33.33 & 1.84 ± 1.28 \\
          & TRPO-L & 309.96 ± 25.77 & 0.93 ± 0.62 & 262.01 ± 14.24 & 1.09 ± 0.13 & 653.86 ± 58.67 & 3.66 ± 0.17 & 14.36 ± 10.01 & 1.32 ± 1.73 \\
    \multicolumn{1}{c}{BC-v3} & SAC-L & 253.25 ± 1.76 & 0.14 ± 0.12 & 0.06 ± 2.88 & 3.57 ± 0.06 & 855.29 ± 0.85 & 3.12 ± 0.04 & -10.89 ± 44.52 & 3.14 ± 1.96 \\
          & DDPG-L & 395.23 ± 71.12 & 1.04 ± 0.60 & 354.78 ± 10.97 & 1.04 ± 0.15 & 936.82 ± 83.08 & 3.06 ± 0.04 & 363.11 ± 14.93 & 0.93 ± 0.13 \\
\cmidrule{2-10}          & Average & \cellcolor[rgb]{ .851,  .851,  .851}\textbf{293.11} & \cellcolor[rgb]{ .851,  .851,  .851}\textbf{0.77} & 164.35 & 1.68  & \cellcolor[rgb]{ .851,  .851,  .851}503.8 & \cellcolor[rgb]{ .851,  .851,  .851}2.92 & 98.87 & 1.81 \\
    \midrule
          & PPO-L & 199.42 ± 28.33 & 0.62 ± 0.49 & 17.85 ± 46.46 & 3.13 ± 3.33 & 211.08 ± 16.37 & 1.85 ± 0.49 & -6.86 ± 14.94 & 5.99 ± 1.45 \\
          & TRPO-L & 175.53 ± 36.64 & 0.65 ± 0.89 & 33.83 ± 85.95 & 1.18 ± 0.66 & 220.11 ± 50.05 & 1.99 ± 1.93 & 36.77 ± 98.88 & 1.01 ± 0.53 \\
    \multicolumn{1}{c}{CC-v3} & SAC-L & 199.66 ± 56.06 & 1.18 ± 1.22 & -66.29 ± 36.21 & 5.23 ± 0.98 & 207.12 ± 19.21 & 1.12 ± 0.73 & -12.37 ± 21.65 & 2.69 ± 1.09 \\
          & DDPG-L & 214.97 ± 4.05 & 0.44 ± 0.06 & 103.78 ± 60.52 & 2.20 ± 1.65 & 213.63 ± 8.37 & 0.88 ± 0.44 & 1.07 ± 36.21 & 1.24 ± 0.49 \\
\cmidrule{2-10}          & Average & \cellcolor[rgb]{ .851,  .851,  .851}\textbf{197.38} & \cellcolor[rgb]{ .851,  .851,  .851}\textbf{0.72} & 22.29 & 2.94  & \cellcolor[rgb]{ .851,  .851,  .851}212.99 & \cellcolor[rgb]{ .851,  .851,  .851}1.44 & 4.65  & 2.73 \\
    \midrule
          & PPO-L & 416.34 ± 59.33 & 0.97 ± 0.11 & 257.29 ± 21.35 & 1.54 ± 0.14 & 426.94 ± 61.96 & 1.92± 0.27 & 215.96 ± 125.53 & 1.52 ± 0.57 \\
          & TRPO-L & 554.44 ± 55.20 & 1.05 ± 0.12 & 535.91 ± 57.58 & 2.01 ± 0.10 & 539.57 ± 21.83 & 2.30 ± 0.82 & 525.27 ± 56.78 & 2.11 ± 0.14 \\
    \multicolumn{1}{c}{DC-v3} & SAC-L & 590.60 ± 224.05 & 1.00 ± 0.12 & 114.94 ± 80.11 & 1.48 ± 0.79 & 728.02 ± 217.83 & 3.57 ± 1.42 & 194.09 ± 117.81 & 1.49 ± 1.16 \\
          & DDPG-L & 643.25 ± 77.19 & 1.01 ± 0.04 & 338.91 ± 30.98 & 1.98 ± 0.48 & 839.36 ± 52.66 & 4.43 ± 1.42 & 322.05 ± 86.86 & 1.68 ± 0.46 \\
\cmidrule{2-10}          & Average & \cellcolor[rgb]{ .851,  .851,  .851}\textbf{551.15} & \cellcolor[rgb]{ .851,  .851,  .851}\textbf{1.01} & 311.76 & 1.75  & \cellcolor[rgb]{ .851,  .851,  .851}633.47 & \cellcolor[rgb]{ .851,  .851,  .851}3.06 & \textbf{314.34} & \textbf{1.70} \\
    \bottomrule
    \end{tabular}%
}
\caption{\small 
Evaluation results of the Bullet-safety-gym tasks. The cost threshold is 1.
$\uparrow$ / $\downarrow$: the higher/lower, the better.
Each value is averaged over 20 episodes and 5 seeds. \colorbox{MyBlue}{Shade}: the two most rewarding agents, \textbf{bold}:  all the safe agents (cost-N $\leq 1$) or two safest agents if none is absolutely constraint-satisfactory.  
}
\vspace{-2pt} 
\label{tab:exp-main}
\end{table}

\begin{table}[ht]
\centering
\small
\renewcommand{\arraystretch}{1.1}
\resizebox{1.\linewidth}{!}{
\begin{tabular}{cccccccccc}
    \toprule
    \multirow{2}[4]{*}{Env} & \multirow{2}[4]{*}{Method} & \multicolumn{2}{c}{GradS (ours)} & \multicolumn{2}{c}{Vanilla} & \multicolumn{2}{c}{CRPO} & \multicolumn{2}{c}{Min-Max} \\
\cmidrule{3-10}          &       & Reward $\uparrow$ & Cost-N $\downarrow$ & Reward $\uparrow$ & Cost-N $\downarrow$ & Reward $\uparrow$ & Cost-N $\downarrow$ & Reward $\uparrow$ & Cost-N $\downarrow$ \\
    \midrule
    \ PG-v2 \  & \ PPO-L \ & 16.74 ± 2.05 & 0.84 ± 0.46 & 1.54 ± 1.16 & 1.03 ± 0.29 & 18.27 ± 6.13 & 1.75 ± 0.70 & 6.76 ± 6.39 & 1.39 ± 0.59 \\
    CG-v2  & PPO-L & 30.57 ± 1.77 & 1.11 ± 0.34 & 0.18 ± 0.41 & 1.12 ± 0.64 & 31.35 ± 1.32 & 1.03 ± 0.12 & 2.65 ± 4.67 & 1.20 ± 0.76 \\
    \bottomrule
    \multicolumn{2}{c}{Average} & \cellcolor[rgb]{ .851,  .851,  .851}\textbf{23.66} & \cellcolor[rgb]{ .851,  .851,  .851}\textbf{0.98} & 1.72  & 1.08  & \cellcolor[rgb]{ .851,  .851,  .851}24.81 & \cellcolor[rgb]{ .851,  .851,  .851}1.39 & 4.71  & 1.30 \\
    \midrule
    PG-v3  & PPO-L & 18.09 ± 2.74 & 1.04 ± 0.81 & 7.26 ± 7.87 & 0.85 ± 0.32 & 19.11 ± 2.46 & 1.68 ± 0.42 & 1.35 ± 3.98 & 1.19 ± 1.54 \\
    CG-v3  & PPO-L & 2.22 ± 5.20 & 0.98 ± 0.16 & -2.22 ± 5.20 & 1.28 ± 0.45 & 9.75 ± 5.39 & 1.93 ± 0.33 & -1.15 ± 1.89 & 1.63 ± 1.01 \\
    \bottomrule
    \multicolumn{2}{c}{Average} & \cellcolor[rgb]{ .851,  .851,  .851}\textbf{10.16} & \cellcolor[rgb]{ .851,  .851,  .851}\textbf{1.01} & \textbf{2.52} & \textbf{1.07} & \cellcolor[rgb]{ .851,  .851,  .851}14.43 & \cellcolor[rgb]{ .851,  .851,  .851}1.81 & 0.10  & 1.16 \\
    \bottomrule
    \end{tabular}%
}
\caption{\small Evaluation results of the Safety-gymnasium tasks. The cost threshold is 1.
$\uparrow$ / $\downarrow$: the higher/lower, the better.
Each value is averaged over 20 episodes and 5 seeds. \colorbox{MyBlue}{Shade}: the two most rewarding agents, \textbf{bold}:  all the safe agents (cost-N $\leq 1$) or two safest agents if none is absolutely constraint-satisfactory. We selected PPO-Lag for the base safe RL algorithm since the original single-cost envs are already challenging for others such as SAC-Lag as reported by \cite{liu2023datasets}. 
}
\vspace{-10pt} 
\label{tab:exp-main-safety-gymnasium}
\end{table}

\vspace{-8pt} 
\subsection{Challenges for MC Safe RL}
The performance of the baseline method \texttt{Vanilla} is summarized in Table. \ref{tab:exp-main}, \ref{tab:exp-main-safety-gymnasium} and Figure. \ref{fig: Scalability analysis}. It is evident that in ``\texttt{-v2}'' settings, \texttt{Vanilla} struggles to learn a rewarding policy 
% within a reasonable number of epochs 
due to the over-conservativeness caused by \textit{redundant} constraints. In ``\texttt{-v3}'' settings, \texttt{Vanilla} encounters difficulties in exploration induced by the \textit{conflicting} constraints, ultimately leading to the failure to learn a reasonable policy. This observation highlights the challenges posed by MC settings for safe RL, as (1) \textit{redundant} constraints contribute to over-conservativeness in the policy update, since the agent would overestimate the effort to ensure safety, and (2) \textit{conflicting} constraints restrict the exploration capabilities of online safe RL algorithms, causing the policy to converge to a local optimum around the initial points, resulting in the agent getting stuck due to the dominating gradients of conflicting constraints if it deviates from this point. The unsatisfactory reward and safety performance of \texttt{Vanilla} methods in MC safe RL settings underscore the importance of exploring efficient MC safe RL algorithms.

% The performance of baseline method \texttt{Vanilla} is summarized in Table. \ref{tab:exp-main} and Figure. \ref{fig: Scalability analysis}. We can observe that in redundant constraint settings, it is hard for \texttt{Vanilla} to learn a rewarding policy within certain epochs, and in conflicting constraint settings, \texttt{Vanilla} can not efficiently explore, thus fail to learn a reasonable policy. This indicates that redundant constraints and conflicting constraints are challenging for MC-safe RL because (1) redundant constraints lead to over-conservativeness in the policy update, thus reducing the rewards of the agents. (2) conflicting constraints confine the exploration of online safe RL algorithms, i.e., it will make the policy converge to a local optimal around the initial points, and the agent will get stuck because of the dominating gradients of conflicting gradients if it deviates from this point. \textbf{The poor reward and safe performance for vanilla methods in MC-safe RL settings indicates the necessity of studying efficient MC-safe RL algorithms.} 
\vspace{-8pt} 
\subsection{GradS performance comparison in MC Safe RL}
The performance of GradS and other baseline methods \texttt{CRPO} and \texttt{Min-Max} is also summarized in Table. \ref{tab:exp-main}, \ref{tab:exp-main-safety-gymnasium}, and Figure. \ref{fig: Scalability analysis}. In ``\texttt{-v2}'' settings with \textit{redundant} constraints, \texttt{Min-Max} exhibits strong performance since it consistently penalizes the most violated constraint, thus eliminating the negative impact of \textit{redundant} constraints, and avoids over-conservativeness in the policy update. However, in ``\texttt{-v3}'' settings with \textit{conflicting} constraints, it struggles to achieve a rewarding policy as it becomes trapped in local optima due to the lack of random exploration. Conversely, the \texttt{CRPO} method explores well with a high reward in conflicting settings, benefiting from its stochastic constraint gradient selection and resulted superior exploration capabilities, thereby avoiding entrapment in local optima. Nevertheless, it fails to ensure constraint satisfaction in the presence of \textit{redundant} constraints due to potential imbalances between different types of cost. Specifically, if one type of \textit{redundant} constraints significantly outweighs others in terms of quantity, the \texttt{CRPO} method would disproportionately activate constraints from this type, potentially overlooking other constraints. 
% \textbf{The unsatisfactory reward and constraint satisfaction performance further underscore the necessity of developing efficient and safe approaches for MC-safe RL.}

The proposed GradS method demonstrates strong performance, exhibiting high rewards and small cost violations across both ``\texttt{-v2}'' and ``\texttt{-v3}'' MC-safe RL tasks from both benchmarks. In ``\texttt{-v2}'' tasks involving \textit{redundant} constraints, GradS overcomes issues of over-conservativeness akin to those observed in the \texttt{Vanilla} method by eliminating \textit{redundant} gradients. Furthermore, the elimination of redundant constraints reduces the risk of neglecting minor constraints, a drawback of the \texttt{CRPO}. In ``\texttt{-v3}'' scenarios with \textit{conflicting} constraints, GradS excels in performance with high reward and low cost violation compared to the baseline algorithm as it eliminates the conflicting constraints and enables stochastic constraint gradients to encourage exploration.

\vspace{-8pt} 
\subsection{Scalability analysis}
\begin{figure}[t]
\vspace{-5pt} 
\centering
\includegraphics[width=1\linewidth]{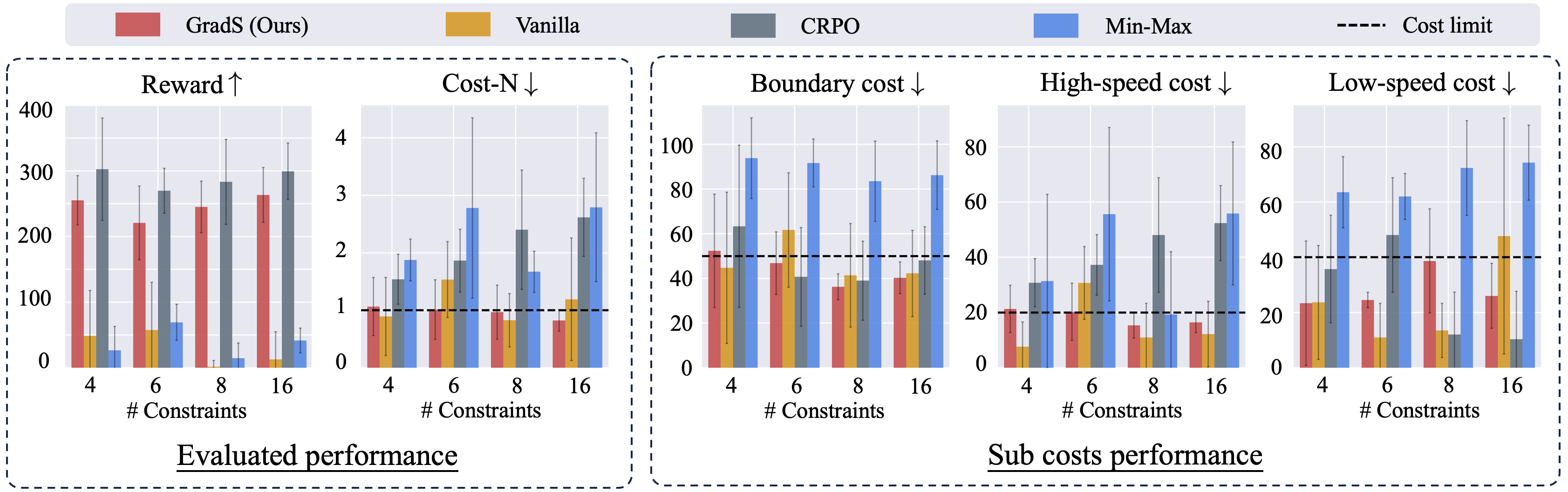}
\caption{\small Scalability analysis: The x-axis in each figure means the constraint number in the tasks. 
The first two figures show the reward and normalized costs, while the remaining three show the representative cost returns. 
The bar charts represent the mean value and the error bars represent the standard deviation. 
All plots are averaged among $5$
random seeds and $10$ trajectories for each seed.
$\uparrow$ / $\downarrow$: the higher/lower, the better.
}
\vspace{-10pt}
\label{fig: Scalability analysis}
\end{figure}

The results of the cost dimension scalability experiment are shown in Figure. \ref{fig: Scalability analysis}. We utilize the Ball-Circle (BC-v3) task to evaluate the algorithms across various constraint quantities. Here we increase the number of costs by creating new constraints with similar velocity thresholds and boundary positions (see appendix for details). It is evident that the baseline methods \texttt{vanilla} and \texttt{Min-max} struggle to learn safe policies, as they encounter difficulties in effectively exploring the action and observation space in the MC tasks. The baseline method \texttt{CRPO} succeeds in learning a rewarding yet unsafe policy, attributed to its lack of ability to manage imbalanced constraints. In contrast, the proposed GradS method demonstrates consistent performance as the number of constraints varies, highlighting the scalability of our approach.

\vspace{-10pt} 
\section{Conclusion}
\vspace{-3pt} 
% Empirical findings have indicated that multi-cost (MC) safe RL poses more challenges compared to single-cost settings \citep{yao2023constraint}. 
In this paper, we proposed a unified framework for Lagrangian-based MC Safe RL algorithms from the standpoint of Multi-Objective Optimization (MOO), and analyze the MC safe RL problem through the lens of constraint types, identifying two challenging MC safe RL settings: \textit{redundant} and \textit{conflicting} constraints. To address these challenges, we propose the constraint gradient shaping (GradS) technique, ensuring compatibility with general Lagrangian-based safe RL algorithms. Our analysis highlights the necessity of developing efficient and effective algorithms for handling multiple costs, shedding light on the critical importance of addressing multi-cost constraints in safe RL settings. 
The extensive experimental results reconfirm that GradS effectively solves the MC safe RL problems in both \textit{redundant} and \textit{conflicting} constraint settings, and is safer, and more rewarding than baseline methods.
By proposing the GradS technique and providing a comprehensive analysis, we hope to contribute to the advancement of safe RL algorithms and their successful implementation in real-world complex and safety-critical environments.
The limitation of this work is the additional computational burden when calculating the gradient similarity. The future work contains the extension to offline safe RL settings.

\newpage

% Acknowledgments---Will not appear in anonymized version
% \acks{We thank a bunch of people.}

\bibliography{sample}

% \clearpage
% \tableofcontents
% \clearpage
% \appendix
% \input{Appendix_tex/Theoritical_analysis}

% \input{Appendix_tex/Implementation_details}

% \input{Appendix_tex/Experiment_supp}

\end{document}